\documentclass[10pt,twocolumn,letterpaper]{article}

\usepackage{iccv}
\usepackage{times}
\usepackage{epsfig}
\usepackage{graphicx}
\usepackage{amsmath}
\usepackage{amssymb}
\usepackage{multirow}
\usepackage{balance}
\usepackage[toc,page]{appendix}

\usepackage[breaklinks=true,bookmarks=false,colorlinks,letterpaper=true]{hyperref}

\iccvfinalcopy 

\def\iccvPaperID{6351} 

\ificcvfinal\fi

\begin{document}

\title{TGRNet: A Table Graph Reconstruction Network \\ for Table Structure Recognition}

\author{
Wenyuan Xue$^*$, 
Baosheng Yu$^\dagger$, 
Wen Wang$^*$, 
Dacheng Tao$^{\ddagger\dagger}$, 
Qingyong Li$^*$\\
\thanks{Qingyong Li and Wen Wang are the corresponding authors.}
Beijing Key Lab of Traffic Data Analysis and Mining, Beijing Jiaotong University, China\\
$^\dagger$The University of Sydney, Australia\\
$^\ddagger$JD Explore Academy, China\\
{\tt\small \{wyxue17,wangwen,liqy\}@bjtu.edu.cn, baosheng.yu@sydney.edu.au,  dacheng.tao@gmail.com}
}

\maketitle
\ificcvfinal\thispagestyle{empty}\fi

\begin{abstract}
 A table arranging data in rows and columns is a very effective data structure, which has been widely used in business and scientific research. Considering large-scale tabular data in online and offline documents, automatic table recognition has attracted increasing attention from the document analysis community. Though human can easily understand the structure of tables, it remains a challenge for machines to understand that, especially due to a variety of different table layouts and styles. Existing methods usually model a table as either the markup sequence or the adjacency matrix between different table cells, failing to address the importance of the logical location of table cells, e.g., a cell is located in the first row and the second column of the table. In this paper, we reformulate the problem of table structure recognition as the table graph reconstruction, and propose an end-to-end trainable table graph reconstruction network (TGRNet) for table structure recognition. Specifically, the proposed method has two main branches, a cell detection branch and a cell logical location branch, to jointly predict the spatial location and the logical location of different cells. Experimental results on three popular table recognition datasets and a new dataset with table graph annotations (TableGraph-350K) demonstrate the effectiveness of the proposed TGRNet for table structure recognition. Code and annotations will be made publicly available at~\url{https://github.com/xuewenyuan/TGRNet}.
\end{abstract}

\section{Introduction}
\begin{figure}
	\centering
	\includegraphics[width=\linewidth]{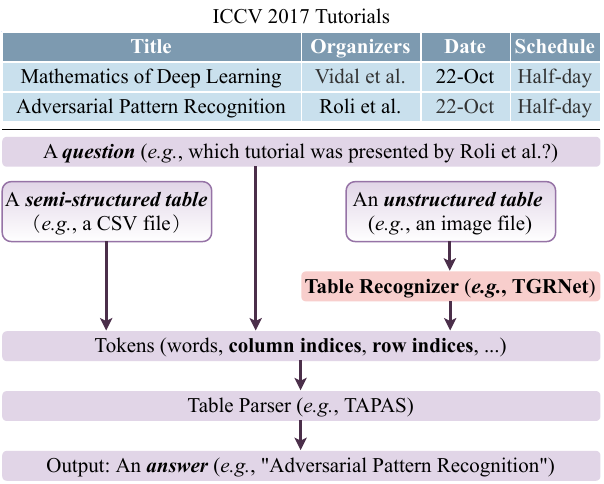}
    \caption{A table parsing example shows the main concern of this paper. Before applying a table parser (\eg, TAPAS~\cite{herzig2020tapas}) to answer a question over the above table, both the table and the question are represented as a sequence of tokens, in which the cell logical location (\ie, the index of column/row) provides the table structure information. Though the cell logical location can be directly acquired from the semi-structured table (\eg, a CSV file), a table structure recognizer is required to obtain such important information from the unstructured table (\eg, an image file).
    }
	\label{qa}
\end{figure}
 
Tabular data have been widely used to help people manage and extract important information in many real-world scenarios, including the analysis of financial documents, air pollution indices, and electronic medical records~\cite{padhi2020tabular,yoon2020vime}. Though human can easily understand tables with different layouts and styles, it remains a great challenge for machines to automatically recognize the structure of various tables. Considering the massive amount of tabular data presented in unstructured formats (\eg, image and PDF files) and that most table analysis methods focus on semi-structured tables (\eg, CSV files)~\cite{herzig2020tapas, yoon2020vime, padhi2020tabular, yin2020tabert}, the community will significantly benefit from an automatic table recognition system, facilitating large-scale tabular data analysis such as table parsing~\cite{herzig2020tapas,yin2020tabert}, patient treatment prediction~\cite{yoon2020vime, zhang2020empowering}, and credit card fraud detection~\cite{padhi2020tabular}.

\begin{figure}
  \centering
  \includegraphics[width=\linewidth]{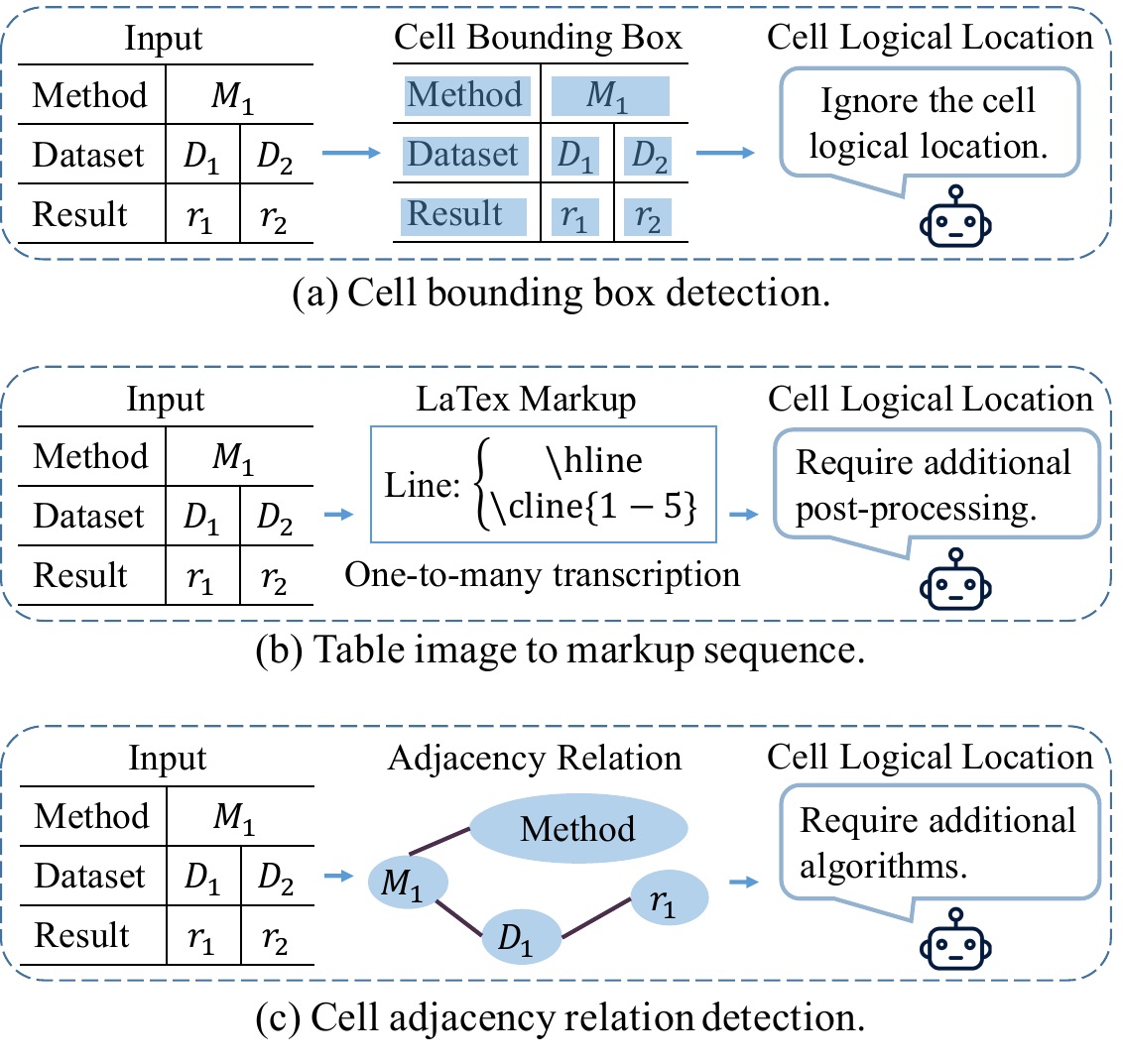}
  \caption{Three types of existing methods for table structure recognition.}
  \label{cmp}
\end{figure}

To understand the structure of different tables, both the cell spatial and logical locations are of great importance in many applications. As a table parsing example shown in Fig.~\ref{qa}, before applying a table parser (\eg, TAPAS~\cite{herzig2020tapas}) to answer a question over a table, both the table and the question are tokenized, and the cell logical location (\ie, the index of column/row) provides the table structure information. If the table is presented as an image instead of structured or semi-structured formats, a table structure recognizer is required to detect the cell spatial location and infer the cell logical location. Existing table structure recognition methods usually utilize rule-based or statistical techniques with hand-crafted features~\cite{oro2009trex,rastan2015texus,shigarov2016configurable}, working well in only constrained settings (\eg, tables with fixed layouts). As shown in Fig.~\ref{cmp}, with the success of deep learning, recent deep learning-based table structure recognition approaches can be divided into three categories: 1) identify cell bounding boxes through visual detection and segmentation methods~\cite{schreiber2017deepdesrt,siddiqui2019deeptabstr,tensmeyer2019deep,siddiqui2019rethinking,khan2019table}; 2) transform a table image into the markup sequence, such as LaTeX and HTML~\cite{li2019tablebank,deng2019challenges}; and 3) explore the adjacency relation between different table cells~\cite{xue2019ReS2TIM,qasim2019rethinking,li2020gfte}. Though the logical location of each cell can be inferred from the adjacency matrix of table cells, additional complex graph optimization algorithms are required. As a result, the importance of the logical location of table cells has been poorly investigated in existing table structure recognition methods.

To further explore the logical relation between different table cells, we introduce a more powerful graph-based table representation, which is referred to as {\em Table Graph}. Specifically, the structure of each table can be represented as a graph: each node indicates a table cell and the edge between two nodes reflects their logical relation on the row and column dimensions, which can be related to their row and column indices. With the proposed table graph, a table cell can be located in the image by the position of its pixels, and its relevant information can be retrieved along the row and column indices. As a result, if a model can reconstruct such a table graph from the given image, it then has a good understanding on the table structure. As the table parsing example shown in Fig.~\ref{qa}, the table structure is represented by the logical location of each cell, which can be inferred from the cell spatial location when the input table is presented as an image. To this end, both the cell spatial and logical locations are important for table structure recognition 
and further table understanding.

In this paper, we formulate the problem of table structure recognition as the table graph reconstruction, which requires the model to jointly predict the cell spatial location and the cell logical location. To address this problem, an end-to-end trainable table reconstruction network (TGRNet) is proposed. Specifically, the proposed method employs a segmentation-based module to detect the cell spatial location, and the cell logical location prediction is solved as an ordinal node classification problem. We evaluate the proposed method on four datasets and experimental results demonstrate the effectiveness of the proposed method for table structure recognition. Considering most table recognition datasets do not provide the cell logical location annotation, we provide the table graph annotations for 350K table images from the TABLE2LATEX-450K dataset \cite{deng2019challenges} as a new benchmark, TableGraph-350K. The contributions of this paper are summarized as follows: 
\begin{itemize}
  \item We reformulate the problem of table structure recognition as the table graph reconstruction, and further propose a table graph reconstruction network to jointly predict the spatial and logical locations of table cells. 
  \item We contribute a new benchmark generated from the TABLE2LATEX-450K dataset with 350K table graph annotations.
\end{itemize}

\section{Related Work}

\subsection{Cell Detection and Segmentation}
\label{related work: cell detection}
Inspired by recent work on semantic segmentation and object detection, some researchers utilized deep learning techniques to detect table cells. DeepDeSRT~\cite{schreiber2017deepdesrt} is a 2-fold system that applies Faster RCNN~\cite{Ren2015} and FCN~\cite{long2015fully} for both table detection and row/column segmentation. Paliwal \emph{et al.}~\cite{paliwal2019tablenet} proposed an end-to-end deep model with one encoder and two decoders for both table and column segmentation. In~\cite{khan2019table,siddiqui2019rethinking,tensmeyer2019deep}, they classify an entire row or column into the cell or non-cell categories instead of the pixel-wise classification. Siddiqui \emph{et al.}~\cite{siddiqui2019deeptabstr} treated the row/column identification as an object detection problem. Prasad \emph{et al.}~\cite{prasad2020cascadetabnet} used a cascade architecture for both table detection and cell segmentation. These work well explore deep vision methods for the cell spatial location detection, while ignore the cell logical location.

\subsection{Table to Markup Sequence}
From the perspective of natural language processing, other researchers tried to convert a table image into the markup sequence (\eg, LaTeX or HTML)~\cite{li2019tablebank,deng2019challenges}. They usually applied an image-to-sequence model that includes an encoder to extract features and a decoder to produce the label sequence. Ideally, the structure of a table can be recognized by parsing the markup sequence. However, the markup sequence contains diversified commands for different styles, which make the table structure can be transcribed into different markup sequences. This one-to-many mapping brings a lot of noise to the ground truth and impedes the model training. Even so, they contribute several large datasets by collecting data from public arXiv articles.

\subsection{Adjacency Relation Detection}
\label{related work: adjacency relation}
As cell detection and segmentation approaches do not consider the cell logical location, some research begin to use the graph structure to explore the relation between different table cells. Generally, these methods can be divided into the edge classification \cite{chi2019complicated,qasim2019rethinking,li2020gfte,xue2019ReS2TIM} and the node classification \cite{lohani2018invoice}. Edge classification methods identify whether two different candidate cells belong to the same cell, row, or column. Node classification methods try to predict the category (\eg, ``date'' or ``price'') of a candidate cell on a specific domain. Xue \emph{et al.} \cite{xue2019ReS2TIM} combined these two kinds of methods. They utilized the neighbor relation among cells to infer the cell logical location. However, most of these methods only explore the cell adjacency relation. When searching information on a table, complex graph optimization algorithms are required to infer the global logical relation from the pair-wise cell adjacency relation.

\begin{figure}[t]
    \centering
    \includegraphics[width=\linewidth]{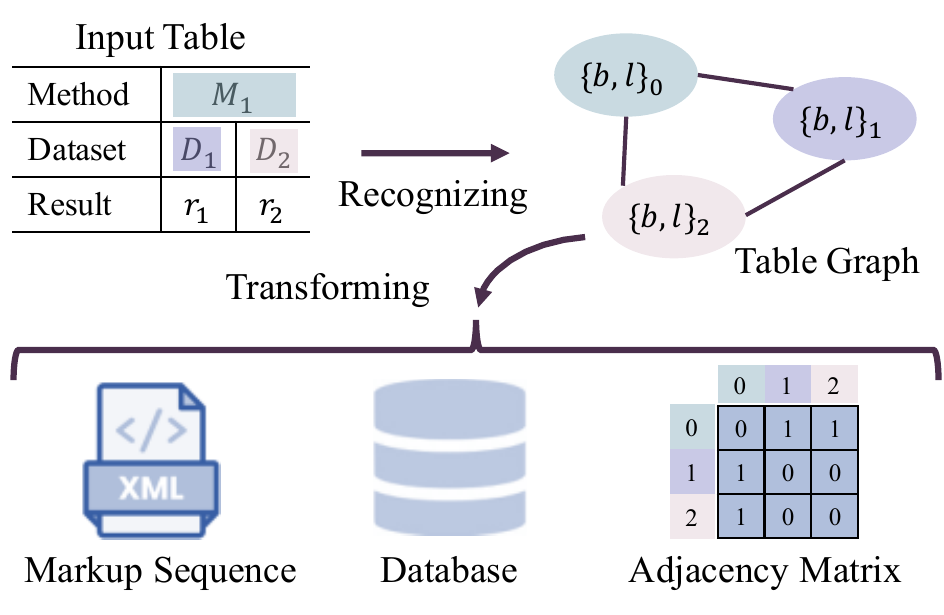}
    \caption{Data transformation from a table graph to other data formats. Three table cells are visualized for simplicity. The adjacency matrix indicates whether two cells belong to the same column.} 
	\label{data-convert}
\end{figure}

\section{Problem Formulation}
In this section, we introduce the formulation of the proposed table graph representation. A table graph is the structured representation of a table in an image, which can be defined as $G = \left(V, A\right)$, where each node in $V$ indicates a table cell and $A$ is the adjacency matrix. Generally, each element in $A$ represents the relation of two different nodes. However, for a table, the logical relation among cells can be represented by their logical indices. Therefore, for each cell or node $v_i$, we denote $b_i$ and $l_i$ as its two attributes that indicate its spatial and logical locations, respectively, 
\begin{gather}
  b_i = (b_i^x, b_i^y, b_i^w, b_i^h), \\
  l_i = ({row}_i^{start}, {row}_i^{end}, {col}_i^{start}, {col}_i^{end}),
\end{gather}
where $(b_{i}^{x}, b_{i}^{y})$, $b_{i}^{w}$, and $b_{i}^{h}$ indicate the centre point, width, and height of the bounding box of $v_{i}$, respectively, and the logical location $l_i$ is composed of its four logical indices, \ie, the start-row, end-row, start-column, and end-column. Because table cells arrange in a two-dimensional space, the adjacency matrix $A$ can be represented based on the Euclidean distance between two different nodes. In this way, we can employ the popular graph convolutional network to learn the graph representation and predict the cell logical location through a node classification way. 

Given a table image, the purpose of table structure recognition then is to reconstruct such a table graph, on which we can locate a cell by its spatial location and retrieve the relevant information according to its logical location, just like that we edit formulae in Microsoft Excel. Different from most existing methods, the table graph can be seen as the metadata of a table structure. Combined with other techniques, the table graph can be transformed into multiple representations in different scenarios. As shown in Fig.~\ref{data-convert}, with the logical location, we can directly build an adjacency matrix to represent the neighbor relation or the same-row and same-column relation between two cells. Followed by an Optical Character Recognition (OCR) engine, the table graph can be transcribed into an XML file or even a database format. On the contrary, though the pair-wise relation between cells or the markup sequence may be suitable in some scenarios, it usually fails to generalize because a complex algorithm is required to infer the global structure on a table.
\begin{figure*}[t]
	\centering
  \includegraphics[width=\linewidth]{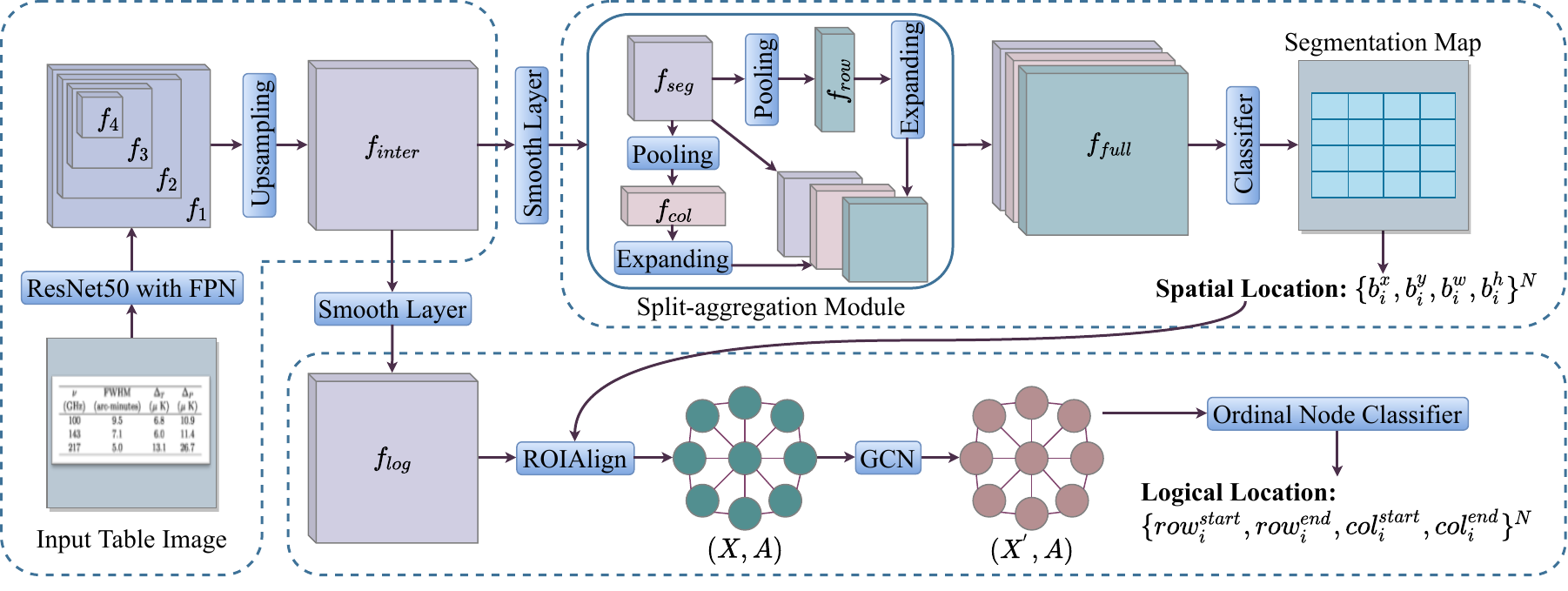}
   \caption{The main framework of the proposed table graph reconstruction network or TGRNet.}
	\label{arch}
\end{figure*}

\section{Method}
In this section, we first describe the main framework of the proposed TGRNet for table structure recognition. We then introduce two main components of TGRNet: cell spatial location detection and cell logical location prediction.

In order to achieve the purpose of reconstructing the table graph from a table image, we devise a table graph reconstruction network (TGRNet) and the main framework is shown in Fig.~\ref{arch}. Specifically, we first employ a backbone network, \eg, ResNet-50~\cite{he2016deep} with FPN~\cite{lin2017feature}, to extract multi-scale feature representations from the input table image. We then jointly perform cell spatial location detection and logical location prediction via two separate head branches in a multi-task manner. For cell spatial location, we first utilize the widely used segmentation-based method to generate a cell segmentation map, from which cells are then detected by computing bounding boxes of connected components. For cell logical location, we apply a graph convolutional network (GCN)~\cite{KipfW17} to learning the table graph representation and solve it as an ordinal node classification problem. In addition, we combine a typical ordinal regression loss with the focal loss~\cite{lin2017focal} as the objective function to address the imbalance problem in cell logical location prediction. We introduce details of cell spatial and logical locations in Section~\ref{sec:method:spatial} and~\ref{sec:method:logical}, respectively.

\subsection{Cell Spatial Location Detection}
\label{sec:method:spatial}

Recently, segmentation-based methods have been popular in table cell detection due to its statistical significance along table rows and columns~\cite{schreiber2017deepdesrt,siddiqui2019deeptabstr,siddiqui2019rethinking}. Therefore, we use a segmentation-based module for the cell spatial location branch to detect the bounding boxes of table cells as follows.

Let $ I \in \mathbb{R}^{3 \times H \times W} $ denote an input table image, where $ H $ and $ W $ indicate the height and width of the input image, respectively. We use four feature maps, $f_1$, $f_2$, $f_3$, and $f_4$, from the ResNet-50 with the stride $s=4,~8,~16,~32$ to construct the feature pyramid~\cite{lin2017feature}: 
\begin{equation}
    f_{inter} = \mathcal{U}_{\times4}(\mathcal{C}(f_1, \mathcal{U}_{\times2}(f_2), \mathcal{U}_{\times4}(f_3), \mathcal{U}_{\times8}(f_4))),
\end{equation}
where the $\mu$ in $\mathcal{U}_{\times \mu}(\cdot)$ is the upsampling scale, $\mathcal{C}(\cdot)$ indicates the channel-wise concatenation operation, all feature maps $f_1, f_2, f_3$, and $f_4$ are converted to $256$ channels via smooth layers. We then have the multi-scale feature representation $f_{inter} \in \mathcal{R}^{1024 \times H \times W}$ as the input of the cell spatial location branch.

To reduce the computational complexity, the cell spatial location branch first reduces the input channels from 1024 to 256 using a $1\times1$ convolutional layer, \ie, $f_{seg} \in \mathbb{R}^{256 \times H \times W}$. Considering that tabular data are arranged in rows and columns, we further introduce a split-aggregation module to utilize the statistical information of row- and column-wise representations. Specifically, the row- and column-wise features, $f_{row} \in \mathbb{R}^{256 \times H}$ and $f_{col} \in \mathbb{R}^{256 \times W}$, are obtained by using the $1 \times W$ and $H \times 1$ average pooling layers, respectively. The row- and column-wise feature representations $f_{row}$ and $f_{col}$ are then expanded to concatenate with the pixel-wise features, \ie,
\begin{equation}
  f_{full} = \mathcal{C}(f_{row}, f_{col}, f_{seg}),
\end{equation}
which is then used to obtain the segmentation map $\hat{y}_{full} \in \mathbb{R}^{K \times H \times W}$, where $K$ indicates the number of classes (\ie, ``background'', ``cell'', and ``boundary''). During the training stage, besides $\hat{y}_{full}$, we also use $f_{row}$ and $f_{col}$ to predict the row- and column-wise segmentation maps, $\hat{y}_{row} \in \mathbb{R}^{K \times H}$ and $\hat{y}_{col} \in \mathbb{R}^{K \times W}$, which can be seen as a kind of statistical regularization. During the testing stage, only $\hat{y}_{full}$ is used to acquire the cell spatial locations by computing the minimum rectangular bounding box for each connected component on the segmentation map.

\subsection{Cell Logical Location Prediction}
\label{sec:method:logical}

We choose candidate table cells from the detected cell bounding boxes to initialize the table graph $G$. A graph convolutional network is then used to learn effective graph representations. Considering that the prediction of logical indices can be seen as a classification with ranking problem, we thus formulate the problem of logical location prediction as an ordinal node classification problem. 

We introduce the construction of the table graph $G$ in detail as follows. During training, the node set $V$ consists of all candidate table cells that have the intersection-over-union (IoU) overlaps with the ground truth table cells larger than 0.5. The feature of the $i$-th node $x_i$ contains two parts: 1) the spatial feature $x^s_{i} \in \mathbb{R}^{256}$ extracted from the spatial location branch; and 2) $x^v_{i} \in \mathbb{R}^{1024}$ obtained by the RoIAlign operation \cite{he2017mask} with the $2\times2$ output size from the smoothed multi-scale feature representation $f_{log} \in \mathbb{R}^{256 \times H \times W}$ according to the bounding box $b_i$ of the $i$-th table cell. We then have the node feature $x_i = \mathcal{C}(x^v_{i}, x^s_{i})$. For the adjacency matrix $A$, each element $a_{i,j}$ indicates the undirected edge between the $i$-th and $j$-th nodes ($i \neq j$). In order to further explore the spatial relation of cells from the row and column dimensions, $a_{i,j}$ is defined as a pair $\{a_{i,j}^{row}, a_{i,j}^{col}\}$, which is calculated based on the Euclidean distance between two nodes:
\begin{equation}
	\begin{cases}
		  a_{i,j}^{row} &= \exp\{-(\frac{b_{i}^{y}-b_{j}^{y}}{H}\cdot\alpha)^2\}, \\
    	a_{i,j}^{col} &= \exp\{-(\frac{b_{i}^{x}-b_{j}^{x}}{W}\cdot\alpha)^2\},
  \end{cases}
  \label{alpha}
\end{equation}
where the adjustment factor $\alpha$ should increase with the growing number of rows or columns, which assigns a large weight to the edges of neighbor cells. After the initialization of the table graph, we then apply a GCN for message passing, 
\begin{equation}
	X^{'} = \rm{ReLU}(\rm{GCN}(\emph{X}, \emph{A})),
\end{equation}
where $X$ and $X^{'}$ indicate the input and output node feature matrixes, respectively. Because the GCN architecture does not support multi-dimensional representations for edges, in our implementation, we use a pair of parallel GCNs to update the node feature matrix for row and column indices prediction, respectively.

With the table graph representation $X^{'}$, we then use an ordinal classifier to predict the cell logical location as follows. Considering that the logical indices of rows and columns in $l_i = ({row}_i^{start}, {row}_i^{end}, {col}_i^{start}, {col}_i^{end})$ are the same problem, for simplicity, we do not distinguish them during the description of the ordinal node classification. Let $r_i \in \{0, 1, ..., T-1\}$ denote the logical index label of the $i$-th node, where $T$ is the total number of rows (or columns). We first convert $r_i$ into a binary label vector $q_i \in \mathbb{R}^{T-1}$ as follows, 
\begin{equation}
	q_{i}^{t} = 
	\begin{cases}
		1, & \text{if} (t < r_i),\\
		0, & \text{otherwise}.
	\end{cases}
\end{equation}
In this way, the logical index prediction is transformed into $T-1$ binary classification sub-problems. The loss function of all 
nodes can be defined as:
\begin{gather}
	\mathcal{L}(X^{'}, \Theta) = -\frac{1}{N}\sum^{N}_{i=1}\Psi(x^{'}_{i},\Theta), \\
	\Psi(x^{'}_{i},\Theta)  = \sum_{t=0}^{r_i-1}\log(p_{i}^{t}) + \sum_{t=r_i}^{T-2}\log(1-p_{i}^{t}), \label{cross-entropy}
\end{gather}
where $N$ is the number of nodes and $p_{i}^{t}$ indicates the predicted probability that $r_i$ is larger than $t$. 
\begin{figure}
	\centering
	\includegraphics[width=\linewidth]{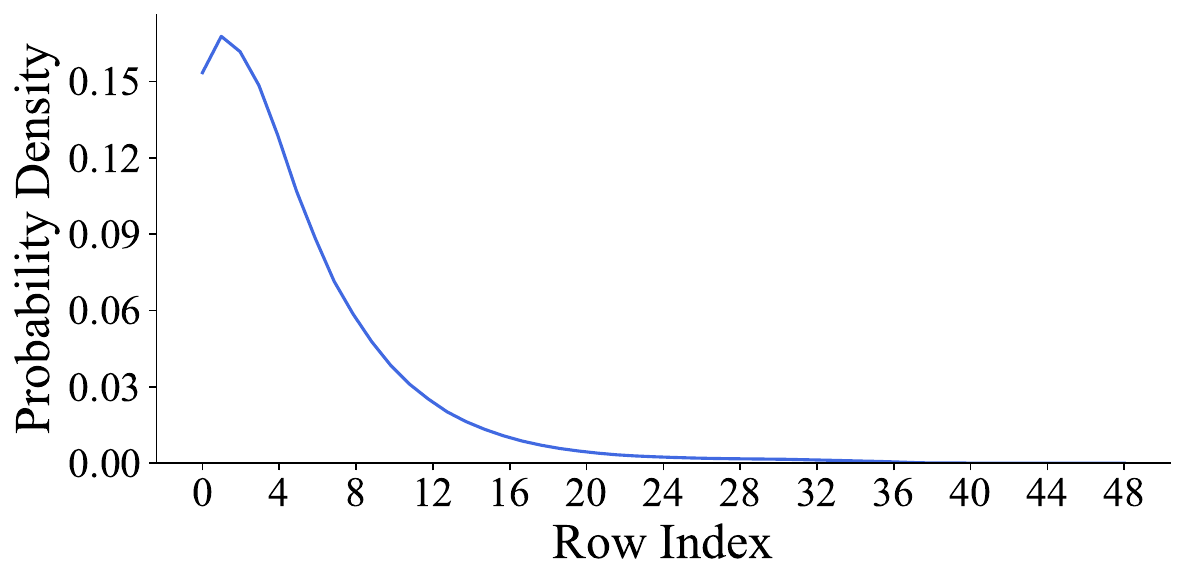}
	\caption{The probability distribution of row indices in TableGraph-350K dataset.}
	\label{row_stat}
\end{figure}

However, the distribution of the cell logical location is usually long-tailed. As shown in 
Fig.~\ref{row_stat}, the number of small indices is much larger than that of large indices. Inspired by the focal loss 
\cite{lin2017focal,lin2020gps}, we address the problem of long-tailed cell logical locations in a similar way, 
\begin{equation}
	\begin{split}
		\Psi(x^{'}_{i},\Theta) =&  \sum_{t=0}^{r_i-1}(1-p_{i}^{t})^{\gamma_{t}}\log(p_{i}^{t}) + \\
		                        & \sum_{t=r_i}^{T-2}(1-p_{i}^{t})^{\gamma_{t}}\log(1-p_{i}^{t})),
	\end{split}
\end{equation}
\begin{equation}
	\gamma_{t} = \min(2, -(1-\lambda_t)^2\log(\lambda_t)+1),
\end{equation}
where $\lambda_t$ is the statistical probability of the logical index $t$ on the training set. The predicted logical index of the $i$-th 
node is the sum of all $T-1$ binary classification results without explicitly ensuring the consistency among the different 
classifiers~\cite{niu2016ordinal}.

\section{TableGraph-350K} 
\label{sec-TG-350K}
Existing table recognition datasets usually have very limited tabular data or lack of the cell logical location labels. Therefore, to build a large-scale benchmark for the table graph reconstruction task, we collect tabular data from the TABLE2LATEX-450K dataset~\cite{deng2019challenges} and provide table graph annotations to generate a new dataset with more than 350K tables, which is referred to as TableGraph-350K\footnote{The details of how we annotate the table graph are presented in supplementary materials.}. For evaluation, we use the same train/val/test split provided by the original dataset. Finally, the new dataset contains 358,767 tables, including 343,988 tables for training, 7,420 tables for validation, and 7,359 tables for testing. The maximum indices of rows and columns are 48 and 27, respectively. 

\section{Experiments}
In this section, we first introduce the datasets, evaluation metrics, and baseline methods. We then present the overall performance of TGRNet for table structure recognition. To demonstrate the robustness of the proposed method and further analyze the limitation of the adjacency relation-based evaluation, we also conduct experiments on incomplete table graphs and challenging historical documents, respectively. Lastly, we perform ablation studies to demonstrate the effectiveness of the main components in TGRNet.

\begin{table*}[t]
  \caption{The overall performance of TGRNet for end-to-end table graph reconstruction.}
  \label{end-2-end}
  \small
  \centering
  \begin{tabular}{l|c|c|c|c|c|c|c|c|c}
    \hline
     \multirow{2}{*}{Dataset} & \multicolumn{3}{|c|}{Cell Spatial Location} & \multicolumn{5}{|c|}{Cell Logical Location} & \multirow{2}{*}{$F_{\beta=0.5}$} \\
    \cline{2-9}
                                         & \emph{P} & \emph{R} & \emph{H} & $A_{rowSt}$ & $A_{rowEd}$ & $A_{colSt}$ & $A_{colEd}$ & $A_{all}$ & \\
    \hline
     ICDAR13-Table & 0.682 & 0.652 & 0.667 & 0.445 & 0.445 & 0.700 & 0.692 & 0.275 & 0.519 \\
    \hline
     TableGraph-24K                      & 0.916 & 0.895 & 0.906 & 0.917 & 0.916 & 0.919 & 0.923 & 0.832 & 0.890 \\
    \hline
  \end{tabular}
\end{table*}

\begin{table*}[t]
 \caption{Experimental results with the metric $A_{all}$ for robustness analysis.}
 \label{incomplete-graphs}
 \small
 \centering
 \begin{tabular}{l|c|c|c|c|c|c}
   \hline
   \multirow{2}{*}{Method}  & \multicolumn{3}{|c|}{CMDD} & \multicolumn{3}{|c}{ICDAR13-Table} \\
   \cline{2-7}
                            & 100\% cells & 90\% cells & 80\% cells & 100\% cells & 90\% cells & 80\% cells \\
   \hline
   ReS2TIM \cite{xue2019ReS2TIM} & 0.999 & 0.941 & 0.705 & 0.174 & 0.137 & 0.124 \\
   \hline
   TGRNet                       & 0.995 & 0.955 & 0.857 & 0.334 & 0.314 & 0.314 \\
   \hline
 \end{tabular}
\end{table*}

\begin{table*}[t]
  \caption{Comparable results for cell logical location and adjacency relation on ICDAR19-cTDaR (TrackB1).}
  \small
  \centering
  \label{icdar19-tgr}
  \begin{tabular}{l|c|c|c|c|c|c|c|c|c}
    \hline
    \multirow{2}{*}{Method} & \multicolumn{3}{|c|}{Cell Spatial Location} & \multicolumn{5}{|c|}{Cell Logical Location} & \multirow{2}{*}{\emph{WAF}}\\
    \cline{2-9}
              & \emph{P} & \emph{R} & \emph{H} & $A_{rowSt}$ & $A_{rowEd}$ & $A_{colSt}$ & $A_{colEd}$ & $A_{all}$ &  \\
    \hline
    ReS2TIM \cite{xue2019ReS2TIM}  & - & - & - & 0.230 & 0.223 & 0.562 & 0.492 & 0.138 & 0.481 \\
    \hline
    TGRNet   & 0.860 & 0.798 & 0.828 & 0.551 & 0.546 & 0.542 & 0.534 & 0.267 & 0.283\\
    \hline
  \end{tabular}
\end{table*}

\subsection{Datasets}
\begin{itemize}
  \item \textbf{TableGraph-24K}. Considering the computational complexity for training a variety of models, we also randomly choose a subset of TableGraph-350K for the academic community, which is referred to as TableGraph-24K. Specifically, this subset contains 20,000 tables for training, 2,000 tables for validation, and 2,000 tables for testing. The maximum indices of rows and columns are 37 and 21, respectively.
  \item \textbf{CMDD} \cite{xue2018table}. This is a medical laboratory report dataset including 476 tables (372 for training and 104 for test). The maximum indices of rows and columns are 24 and 5, respectively. There is no spanning cell and empty cells without texts are not annotated.
  \item \textbf{ICDAR13-Table} \cite{gobel2013icdar}. This dataset consists of 156 tables with spanning cells and other various styles. Empty cells without texts are not annotated. The maximum indices of rows and columns are 57 and 12, respectively. Because the original dataset does not specify the training and test sets, we use half tables for training and others for test following the setting in \cite{xue2019ReS2TIM}.
  \item \textbf{ICDAR19-cTDaR (TrackB1)} \cite{gao2019icdar}. This dataset contains 750 pages from archival historical documents, from which 881 tables are extracted (679 for training and 202 for test). The maximum indices of rows and columns are 87 and 43, respectively. The largest table in this dataset includes more than 2,000 cells.
\end{itemize}

\subsection{Evaluation Metrics}
For cell spatial location detection, we use the same evaluation metrics with recent methods~\cite{schreiber2017deepdesrt,siddiqui2019deeptabstr,tensmeyer2019deep,siddiqui2019rethinking,paliwal2019tablenet}. The predicted cell boxes are evaluated by using the \emph{Precision (P)}, \emph{Recall (R)}, and \emph{Hmean (H)} with the IoU threshold 0.5. 

For cell logical location prediction, we follow the metrics in \cite{xue2019ReS2TIM} to calculate the accuracy of four logical indices (\ie, the start-row, end-row, start-column, and end-column) based on the detected table cells, which are denoted as $A_{rowSt}$, $A_{rowEd}$, $A_{colSt}$, and $A_{colEd}$, respectively. We also evaluate the overall accuracy $A_{all}$ that all four logical indices are predicted correctly for each detected table cell. For using F-Score, $\beta$ is set to be 0.5 to encourage more candidate cell boxes in practice:
\begin{equation}
   F_{\beta=0.5}=\frac{(1+0.5^2) \cdot H \cdot A_{all}}{0.5^2 \cdot H + A_{all}}.
\end{equation}

In Section \ref{exp:CAR}, we also report experimental results using the cell adjacency relation-based metric, \ie, the weighted average F-Score (\emph{WAF})\footnote{We denote the weighted average F-Score as \emph{WAF} for short instead of \emph{WAvg. F1} in ~\cite{gao2019icdar}.}~\cite{gao2019icdar}. For each table cell, the adjacency relations are generated with its nearest neighbours at four directions, \eg, ``up'', ``down'', ``left'', and ``right''. \emph{WAF} calculates the weighted 
F-Score with $\beta=1$ for the cell adjacency relation under different IoU thresholds:
\begin{equation}
    WAF = \frac{\sum^{4}_{i=1}{IoU}_{i} \cdot F_{\beta=1}@{IoU}_{i}}{\sum^{4}_{i=1}{IoU}_{i}},
\end{equation}
where $IoU = \{0.6, 0.7, 0.8, 0.9\}$.

\subsection{Baseline Methods}
For cell logical location prediction, we compare the proposed method with ReS2TIM~\cite{xue2019ReS2TIM}. To the best of our knowledge, in previous methods, ReS2TIM is the only one that aims to predict the cell logical location and reports the evaluation results. Different from TGRNet, ReS2TIM does not contain a cell spatial location module.

For cell spatial location detection, many methods have emerged recently as introduced in Section~\ref{related work: cell detection}. However, most of them were evaluated under different experimental settings due to the lack of standard benchmarks. For example, the DeepDeSRT~\cite{schreiber2017deepdesrt} model was evaluated on a random subset of ICDAR13-Table. Tensmeyer \emph{et al.}~\cite{tensmeyer2019deep} trained the SPLERGE model on a private dataset and evaluated it on ICDAR13-Table by randomly choosing a subset as well. In the literatures~\cite{siddiqui2019rethinking,siddiqui2019deeptabstr}, methods were evaluated on ICDAR13-Table with table cell boxes instead of the original text-level bounding box annotations. It should be noticed that a cell box is larger than the bounding box of the text within that cell. Therefore, we do not present a comparable experiment with these methods for cell spatial location detection because it is hard to do a fair comparison.

\subsection{Overall Performance}
As an end-to-end solution, we present the overall performance of TGRNet in this subsection. ICDAR13-Table and TableGraph-24K are employed for evaluation. The adjustment factor $\alpha$\footnote{\label{foot:alpha}We present the experiment for $\alpha$ selection in supplementary materials.} 
in Eq. \eqref{alpha} is set to be 3. Each input image is resized into $480\times480$ pixels. Because the cell logical location branch requires the cell bounding boxes from the cell spatial location branch to extract the corresponding features, we adopt a pre-training strategy to accelerate the training process. Specifically, before training the whole model on TableGraph-24K, we pre-train TGRNet for 50 epochs while freezing the cell logical location branch. When training on ICDAR13-Table, the model is initialized with the parameters trained on TableGraph-24K. 

Experimental results are presented in Table \ref{end-2-end}. The proposed model achieves 0.519 and 0.890 for $F_{\beta=0.5}$ on ICDAR13-Table and TableGraph-24K, respectively. Although the model has been pre-trained on TableGraph-24K, for ICDAR13-Table, TGRNet does not perform as well as on TableGraph-24K. The possible reasons may be the insufficient data of ICDAR13-Table (only 78 tables for training and the rest 78 tables for test) and the distributional difference between ICDAR13-Table and TableGraph-24K. 

\begin{table*}[t]
  \caption{Ablation studies for logical location prediction.}
  \label{ablation-studies}
  \small
  \centering
  \begin{tabular}{c|c|c|c|c|c|c|c|c|c|c}
    \hline
     Exp.  & GCN & Ord-Reg & Focal & $A_{rowSt}$ & $A_{rowEd}$ & $A_{colSt}$ & $A_{colEd}$ & $A_{all}$ & \emph{H} & $F_{\beta=0.5}$ \\
    \hline
     1 &  &  &                                & 0.884 & 0.884 & 0.779 & 0.775 & 0.697 & 0.899 & 0.850 \\
    \hline
     2 & \checkmark &  &                      & 0.886 & 0.889 & 0.892 & 0.888 & 0.788 & 0.901 & 0.876 \\
    \hline
     3 & \checkmark & \checkmark &            & 0.903 & 0.909 & 0.919 & 0.922 & 0.824 & 0.898 & 0.882 \\
    \hline
     4 & \checkmark & \checkmark & \checkmark & 0.917 & 0.916 & 0.919 & 0.923 & 0.832 & 0.906 & 0.890 \\
    \hline
  \end{tabular}
\end{table*}

\subsection{Robustness Analysis} 
Graph-based methods involve message passing between adjacent nodes. If some key nodes are missing, the inference on the graph will be affected. For the end-to-end table graph reconstruction, the proposed model first detects cell boxes from the input image and then treats them as nodes to predict their logical locations. However, achieving a perfect cell detection result is not easy. In order to evaluate the robustness of TGRNet on incomplete table graphs, we conduct experiments by randomly removing some nodes from the table graph. CMDD and ICDAR13-Table are used for evaluation. Both two datasets do not contain the annotation of empty cells, which means the ground truth may be an incomplete table graph. Tables in CMDD have two kinds of layouts, and TGRNet can achieve 0.991 \emph{Hmean} and 0.992 $A_{all}$ for the end-to-end table graph reconstruction on this dataset. Because ReS2TIM \cite{xue2019ReS2TIM} does not contain a cell spatial location module, for a fair comparison, we use the ground truth of cell spatial locations instead.

As shown in Table \ref{incomplete-graphs}, ``100\%'', ``90\%'', and ``80\%'' indicate the percentage of the reserved ground truth table cells or nodes in the table graph, and the rest of nodes are randomly removed. The $A_{all}$ results demonstrate that the proposed model can keep a relatively stable performance on incomplete table graphs and outperform ReS2TIM.

\subsection{Logical Location vs Adjacency Relation} \label{exp:CAR}
As introduced in Section~\ref{related work: adjacency relation}, some existing methods model the structure of a table as an adjacency matrix between different cells. We argue the limitation of this representation by conducting experiments on ICDAR19-cTDaR (TrackB1). Tables in this dataset are from archival historical documents, which are more challenging than tables in other datasets due to the deformation caused by shooting angles and the overlapping between strokes and table lines. Moreover, some tables are larger than those in other datasets. The means of image heights and widths are 3,298 and 3,149 pixels, respectively. The maximum number of cells within a table is 2,267. So, it will be hard to distinguish cells when resizing a table image into a small scale before feeding it into the model. Due to these challenges, we make some different settings from previous experiments: 
\begin{itemize}
  \item Considering the limited GPU memory, TGRNet is trained separately for cell spatial and logical locations prediction. In the meantime, the input table image is resized into $800 \times 800$ pixels.
  \item During cell spatial location detection, TGRNet applies the morphological open operation with a $3 \times 3$ kernel on the segmentation map to separate squeezed cells.
  \item When initializing the table graph for cell logical location prediction, only $8 \times N$ edges with higher weights are reserved. The adjustment factor $\alpha$\textsuperscript{\ref{foot:alpha}} is set to be 10.
  \item For ReS2TIM, the proportion of positive and negative neighbor relations is kept to $1:4$ by sampling cells around each target cell.
\end{itemize} 

We present experimental results in Table \ref{icdar19-tgr}. TGRNet achieves 0.828 \emph{Hmean} for cell spatial location detection and 0.267 $A_{all}$ for cell logical location prediction. Because ReS2TIM does not contain a cell detection module, it takes the detected cells from TGRNet as input and results in 0.138 $A_{all}$, which is lower than the result of TGRNet. We then compare these two methods based on the cell adjacency relation metric, \ie, \emph{WAF}. Benefiting from the carefully designed pair relation-based loss, ReS2TIM achieves a good result to predict the cell adjacency relation. However, it fails to predict the cell logical location. This reveals the limitation of the cell adjacency relation when understanding the global table structure. Specifically, it is usually required complex graph optimization algorithms to infer the cell logical location from pair-wise relations. However, the cell adjacency relation can be directly gotten from cell logical indices. Moreover, the close results of TGRNet for $A_{all}$ and \emph{WAF} demonstrate its reliability under different evaluation metrics for table structure recognition.

\subsection{Ablation Studies}
\begin{figure}[t]
  \centering
  \includegraphics[width=\linewidth]{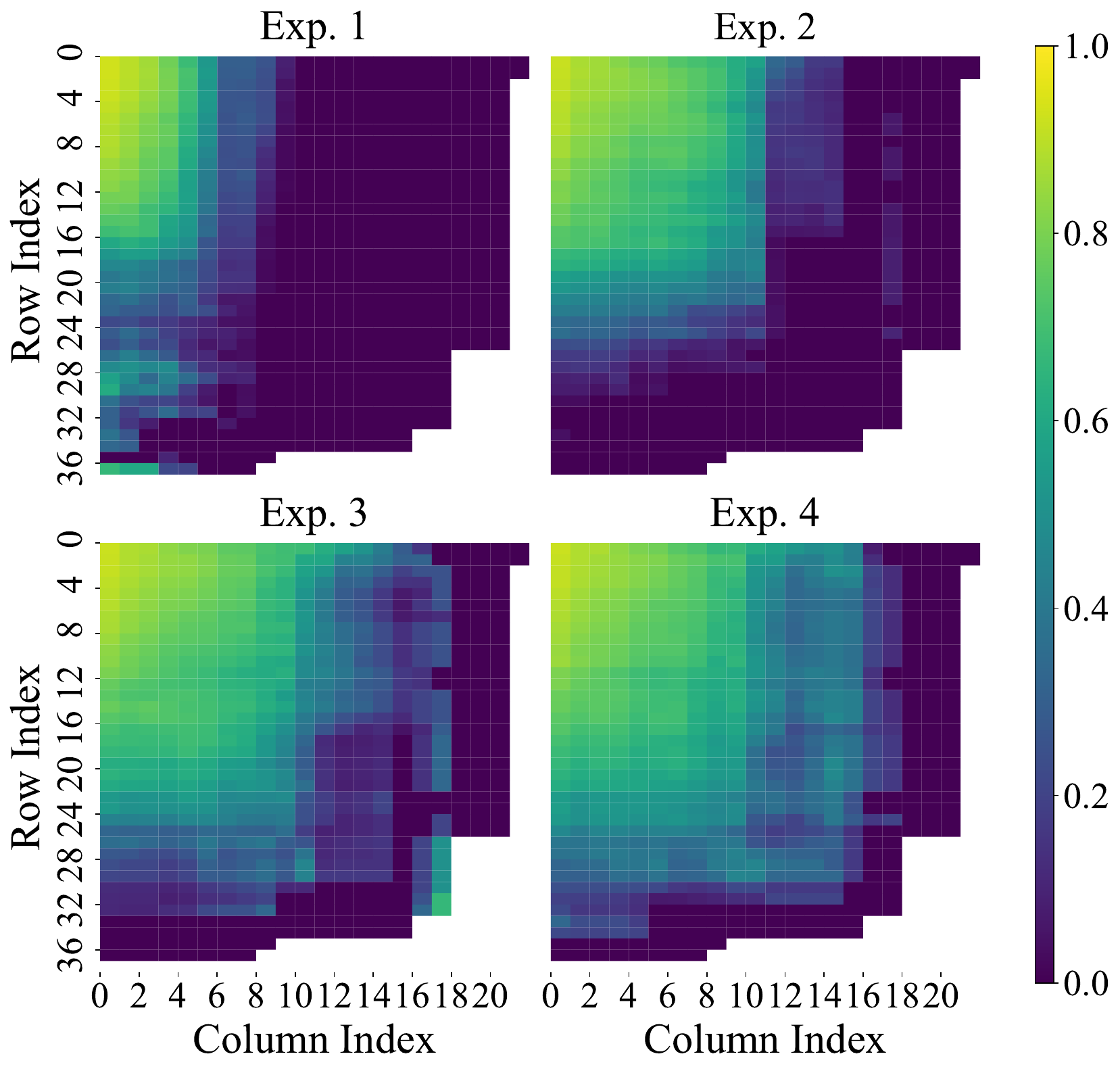}
  \caption{Visualization of the accuracy on each logical location in ablation studies. Locations masked in white mean no cell starts or ends there.}
  \label{vis}
\end{figure}

In this subsection, we isolate the contributions of the main components in TGRNet and present ablation studies on TableGraph-24K. For all experiments shown in Table \ref{ablation-studies}, the backbone network and the cell spatial location branch are unchanged. The baseline of the logical location branch consists of a linear layer with the cross entropy loss, which is denoted as Exp. 1. Then, we add GCN, the ordinal regression loss, and the focal loss to the logical location branch step by step in Exp. 2-4. According to the experimental results, the \emph{Hmean} for cell spatial location detection is around 0.900. GCN and the ordinal regression loss bring significant improvements to the performance, which lead $A_{all}$ from 0.697 to 0.788 and 0.824, respectively. In Fig. \ref{vis}, we visualize the accuracy of each logical location on a heatmap. From the four heatmaps, we can find the baseline model fits well for small logical indices and border indices. When GCN and the ordinal regression loss join in the model, the accuracy of large logical indices increases significantly. The focal loss further improves the performance.

\section{Conclusion}
In this paper, we formulate the problem of table structure recognition as the table graph reconstruction, which requires the model to jointly predict the cell spatial location and the cell logical location. TGRNet is proposed to solve such a problem, which uses a segmentation-based module to detect the cell spatial location and solves the cell logical location prediction as an ordinal node classification problem. 
Experiments on four datasets demonstrate the effectiveness and robustness of TGRNet. Moreover, we contribute a new benchmark generated from TABLE2LATEX-450K dataset 
with 350K table graph annotations.

\section*{Acknowledgments}
This work is supported in part by the National Natural Science Foundation of China under Grant U2034211, 62006017, in part by the Fundamental Research Funds for the Central Universities under Grant 2020JBZD010, and in part by the Beijing Natural Science Foundation under Grant L191016. Dr. Baosheng Yu is supported by ARC project FL-170100117.

{\small
\bibliographystyle{ieee_fullname}
\bibliography{refer.bib}
}
\balance
\clearpage

\appendix

\begin{appendices}
\section{TableGraph-350K Annotation}
\label{appendix:TableGraph-350K}
\begin{figure}[ht]
	\centering
  \includegraphics[width=\linewidth]{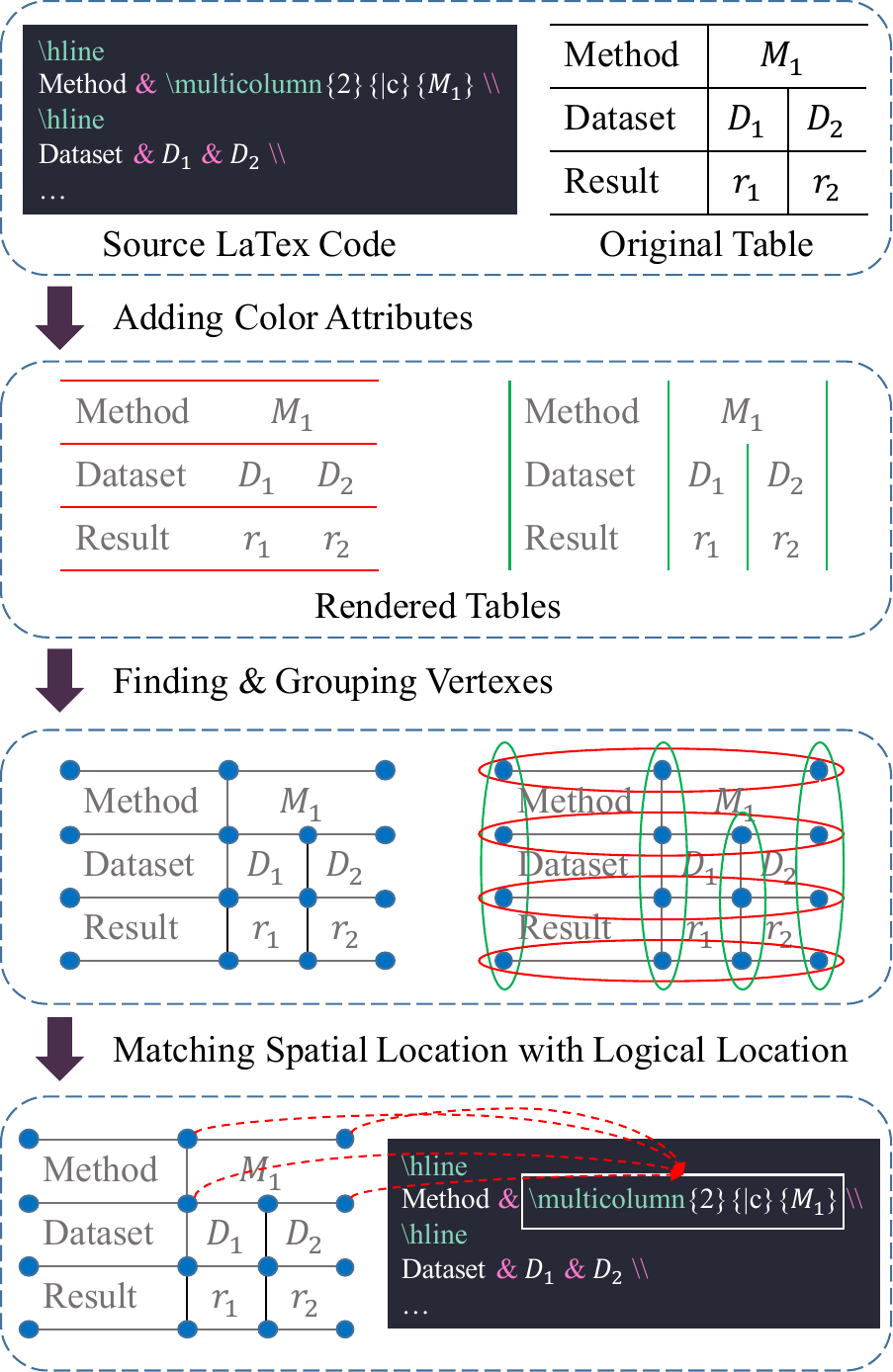}
   \caption{The annotation process of TableGraph-350K.}
	\label{annotation}
\end{figure}

TableGraph-350K is generated from the tabular dataset TABLE2LATEX-450K that 
collects about 450,000 tables from arXiv articles. Each of them is attached with the corresponding rendered 
image and annotated with the LaTeX source code. We present the annotation process in Fig.~\ref{annotation}. During the stage of annotation, 
we add different color attributes to the row and column borderlines in the LaTeX code. The modified LaTeX code can be rendered 
into two table images. One of them has red row borderlines and another has green column borderlines. We identify cell vertexes 
(\ie, the spatial location) by finding pixels that are painted with different colors in these two images. These vertexes are 
further grouped into different rows and columns according to their spatial positions and each vertex also gets its logical location. 
We then match the cell logical location in the LaTeX code with the vertex logical location in the image to complete the table graph 
annotation. As an example shown in Fig.~\ref{annotation}, we can parse LaTeX code to identify that 
the logical location of the ``$M_1$'' cell is $\left(0,0,1,2\right)$, which indicates the start-row, end-row, start-column, and end-column, respectively. 
The left-top and right-bottom vertexes of the ``$M_1$'' cell should have the logical locations of $\left(0,0,1,1\right)$ and $\left(1,1,3,3\right)$, respectively. 
At last, we take the spatial location of the left-top and right-bottom vertexes as the spatial location of the ``$M_1$'' cell.

\section{Hyper-Parameter Selection}
In this section, we show how the adjustment factor $\alpha$ plays a role in the adjacency matrix $A$ and present experiments for the selection of $\alpha$. 

\begin{figure}[ht]
	\centering
  \includegraphics[width=\linewidth]{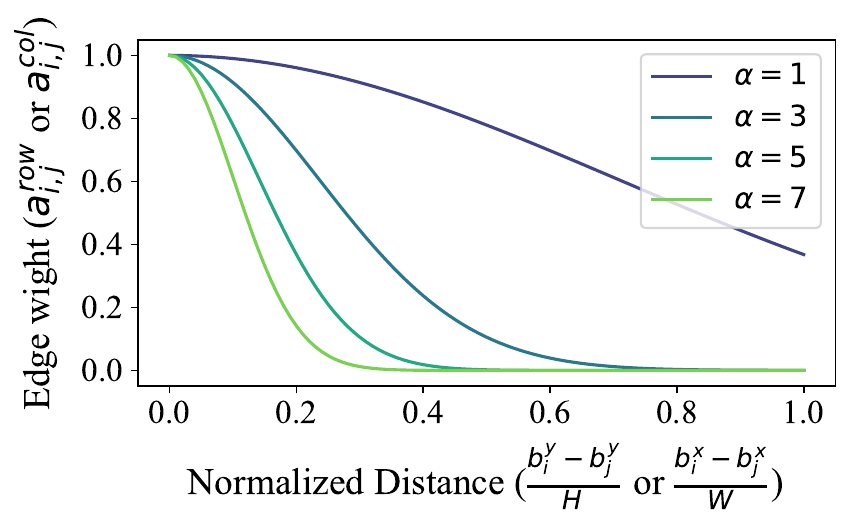}
   \caption{The mapping function from the normalized Euclidean distance (\ie, $\frac{b_{i}^{y}-b_{j}^{y}}{H}$ or 
   $\frac{b_{i}^{x}-b_{j}^{x}}{W}$) to the edge weight (\ie, $a_{i,j}^{row}$ or $a_{i,j}^{col}$) with different values of the adjustment factor $\alpha$.}
	\label{a-alpha}
\end{figure}

\begin{figure}[t]
	\centering
  \includegraphics[width=\linewidth]{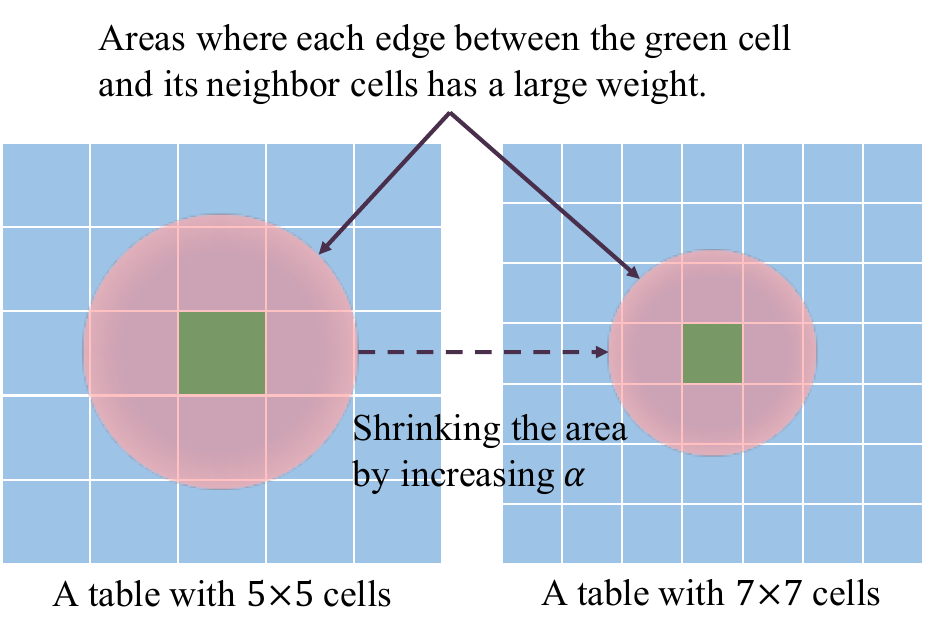}
   \caption{Visualization of how the adjustment factor $\alpha$ plays a role in the adjacency matrix $A$. When the density of cells increases, 
   \eg, the number of cells increases from $5\times5$ to $7\times7$ in the same table area, increasing the value of $\alpha$ can reduce the number of edges that have large weights.}
	\label{tb-alpha}
\end{figure}

During the stage of cell logical location prediction, the graph convolutional network utilizes the adjacency matrix $A$ for message passing between different nodes or table cells. 
However, if the adjacency matrix $A$ connects any different nodes without distinguishing them, \eg, a totally connected graph, 
the graph convolutional network will push representations of adjacent nodes mixed with each other and eventually result in the over-smoothing phenomenon. 
To address this problem, we build the adjacency matrix $A$ by assigning a weight for each edge based on the Euclidean distance of two nodes. As shown in Fig. \ref{a-alpha}, 
if the $i$-th and $j$-th nodes are far away, their edge will be assigned a small weight according to Eq. (5). In this way, messages from faraway nodes 
will be suppressed when the graph convolutional network aggregates information from adjacent nodes. In Fig. \ref{tb-alpha}, we use red circles to represent the areas where 
each edge between the green cell and its neighbor cells has a large weight. When the density of cells increases, \eg, the number of cells increases from $5\times5$ to $7\times7$ in the same table area, we increase $\alpha$ to reduce the number of edges with large weights. 

\begin{figure}[t]
	\centering
  \includegraphics[width=\linewidth]{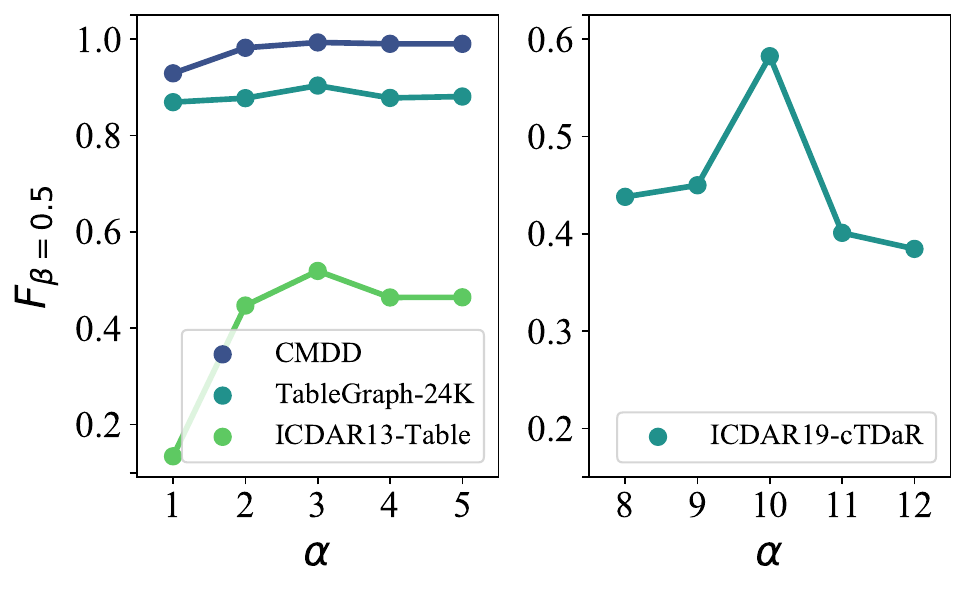}
   \caption{Experimental results for the selection of the adjustment factor $\alpha$.}
	\label{select-alpha}
\end{figure}

As shown in Fig.~\ref{select-alpha}, we present the experimental results of the proposed model or TGRNet with different values of the adjustment factor $\alpha$. For the datasets CMDD, 
TableGraph-24K, and ICDAR2013-Table, their maximum row indices are 24, 37, and 57, respectively, and their maximum column indices are 5, 21, and 12, respectively. 
We change $\alpha$ from 1 to 5 by adding 1 at each time. TGRNet achieves the highest $F_{\beta=0.5}$ on these three datasets when $\alpha = 3$. This demonstrates that the 
adjustment factor $\alpha$ is not a sensitive hyper-parameter. For the dataset ICDAR2019-cTDaR, its maximum indices of rows and columns are 87 and 43, respectively, 
which are larger than the maximum indices in other datasets. When $\alpha = 10$, TGRNet achieves the highest $F_{\beta=0.5}$ on ICDAR2019-cTDaR. Therefore, in previous experiments, 
we choose $\alpha = 10$ and $\alpha = 3$ for ICDAR2019-cTDaR and other datasets, respectively.

\section{Qualitative Comparison}
\begin{figure}[ht]
	\centering
  \includegraphics[width=\linewidth]{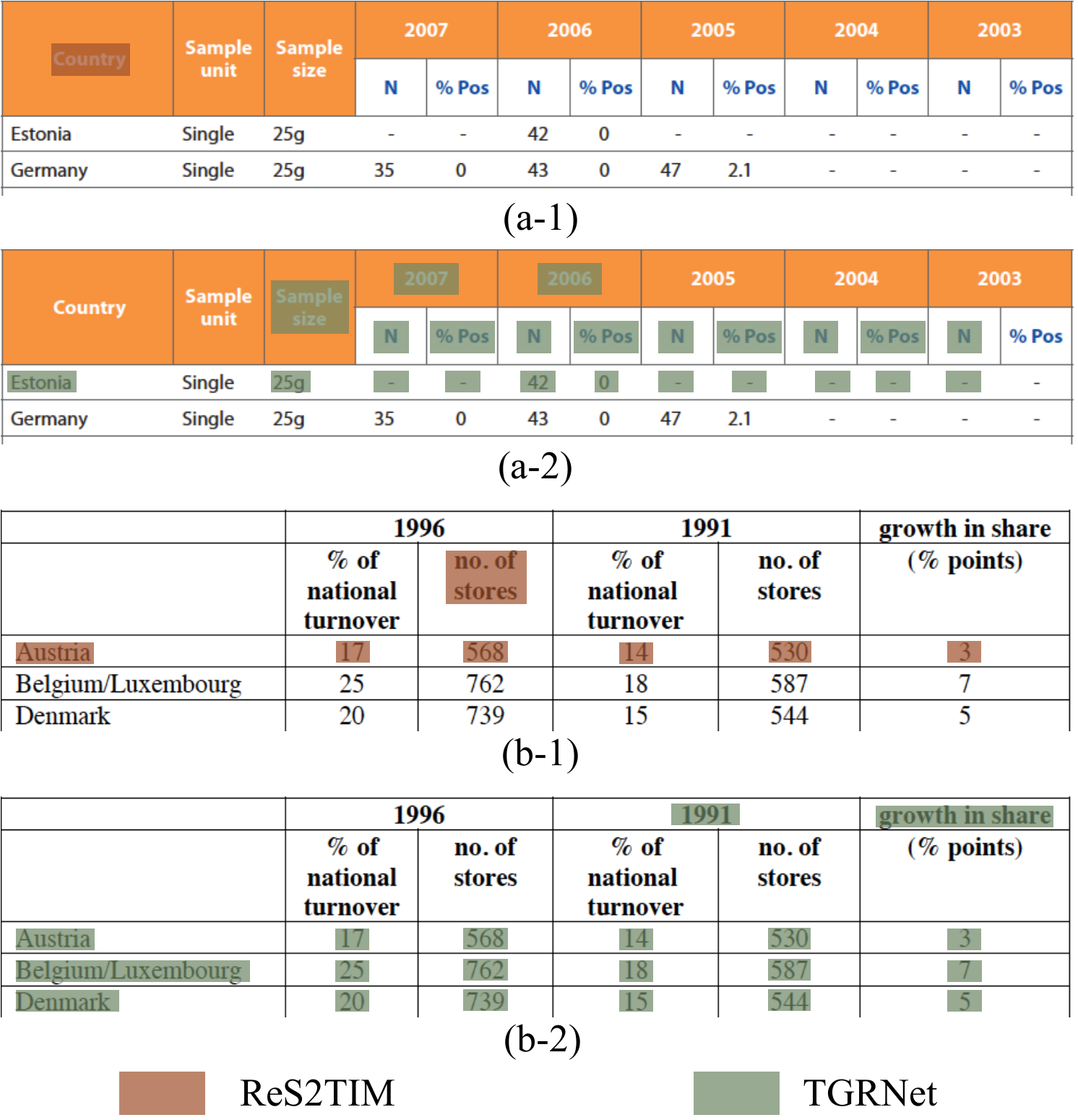}
   \caption{Qualitative comparison. In subfigures (a) and (b), two tables from the ICDAR13-Table dataset are visualized for qualitative comparison, 
   where correct logical location predictions of ReS2TIM and TGRNet are masked with different colors.}
	\label{case}
\end{figure}

In this section, we present visualized results for qualitative comparison. As shown in Fig.~\ref{case}, on the two tables from the ICDAR13-Table dataset, the correct predictions of 
the cell logical location are masked with different colors for ReS2TIM and TGRNet, respectively. Compared with ReS2TIM, TGRNet outputs more correct predictions, especially for 
the cells that span multiple rows or columns. In subfigure (b-2), TGRNet can correctly predict the logical locations of the cells in the last three rows even when it fails for 
all cells in the second row. This demonstrates that TGRNet is less sensitive for missing cells or error accumulation. However, ReS2TIM infers the cell logical location from 
the pair-wise relations, which is susceptible to error accumulation because of the problem of searching paths and the dependence of prior knowledge. 

\end{appendices}
\end{document}